\documentclass{article}
\usepackage{spconf,amsmath,graphicx,hyperref}
\makeatletter
\long\def\@makecaption#1#2{%
  \vskip\dimexpr\topskip-\ht\strutbox\relax
  \setbox\@tempboxa\hbox{\strut #1. #2}%
  \ifdim \wd\@tempboxa >\hsize \strut #1. #2\par \else
    \hbox to\hsize{\hfil\box\@tempboxa\hfil}\fi
  \vskip\belowcaptionskip}
\makeatother
\usepackage{bm}
\usepackage{multirow}

\newcommand{\ret}{\texttt{<ret>}}

\usepackage{fancyhdr}
\fancypagestyle{preprint}{%
  \fancyhf{}%
  \fancyfoot[C]{\footnotesize Preprint version, Under review at ICASSP 2027.}%
}

\title{Context Spanning: A Communication Framework for \\Full-Duplex Speech Models and External LLM Backends}

\name{Seonghyeon Go, Yongwoo Kim, Hyeonjin Cha, Jaeho Shin}
\address{Mindlogic\\ \{seonghyeon.ko, yongwoo, hyeonjin.cha, jaeho.shin\}@mindlogic.ai}
\begin{document}
\maketitle
\thispagestyle{preprint}   
\fontsize{9pt}{11pt}\selectfont
\maketitle

\begin{abstract}
Full-duplex spoken dialogue models can listen and speak simultaneously like the real-time dynamics of human conversation. For natural dialogue, the ability to search for external information in real-time is also an important capability. Many models remain trapped in parametric knowledge, leaving them unable to access real-time information and tool execution. Furthermore, even when Large Language Models (LLM) retrieve information, many duplex speech models process it within a compressed latent space rather than in its raw text form, which can lead to information loss from compression. To address this issue, we propose \textbf{Context Spanning}, a framework for information injection between a full-duplex speech model and an external LLM backend via real-time chunked prefill. The injected frame is encoded in a single forward pass inside the real-time frame budget. It feeds the retrieved information to the speech model as-is, enabling it to reason over the information independently and generate responses. With this approach, our model achieves high performance on Full-Duplex benchmarks and strong results on Question Answering tasks, demonstrating its conversation potential. Context Spanning shows that external information can be injected directly into a duplex speech model, introducing a new simple and powerful mechanism for duplex systems.
\end{abstract}


\section{Introduction}
\label{sec:intro}
    
A full-duplex spoken dialogue model is a system that listens and speaks simultaneously. It enables natural spoken interaction by producing backchannels, generating rapid responses, and yielding the floor when the user interrupts. Moshi\cite{defossez2024moshi} is a foundation model that processes audio from both speakers as parallel streams flowing along the same timeline by integrating listening models, thinking LLMs, and speaking models in a unified system. 
However, to enable truly dynamic and human-like conversation, the ability to retrieve real-time information is important. In LLMs, such tasks have been mainly implemented based on Retrieval Augmented Generation 
(RAG)\cite{lewis2020retrieval}.


Recent studies have also tried to address the application of RAG in duplex spoken dialogue models.  MoshiRAG\cite{chien2026moshirag} demonstrates a framework that communicates asynchronously by a separate backend LLM to gather real-time information. It compresses the retrieved text and adds it onto the user audio token vector to generate responses with external knowledge. Because the information is added as latent representations, it can be distorted before the model reads it. Additionally, injecting all the information may take a long time, depending on the length of the latent vector.

We propose Context Spanning, a framework that addresses this limitation through the direct injection of retrieved information into the speech model. The model can deliver precise values such as time, stock prices, or weather by avoiding the information loss. In addition, because the retrieved information is injected by a single prefill method, it can be done in a very short time budget. Experimental results demonstrate that Context Spanning improves question answering performance while preserving full-duplex conversational abilities. Our main contribution is showing that chunked-prefill can be integrated into the autoregressive architecture of a full-duplex speech model with real-time frame decoding, providing the foundation for our direct context injection framework. We release our full implementation, including training pipelines and evaluation scripts. \footnote{\url{https://github.com/mindlogic-ai/ContextSpanning}}

\begin{figure*}[t]
\centering
\includegraphics[width=0.75\textwidth]
{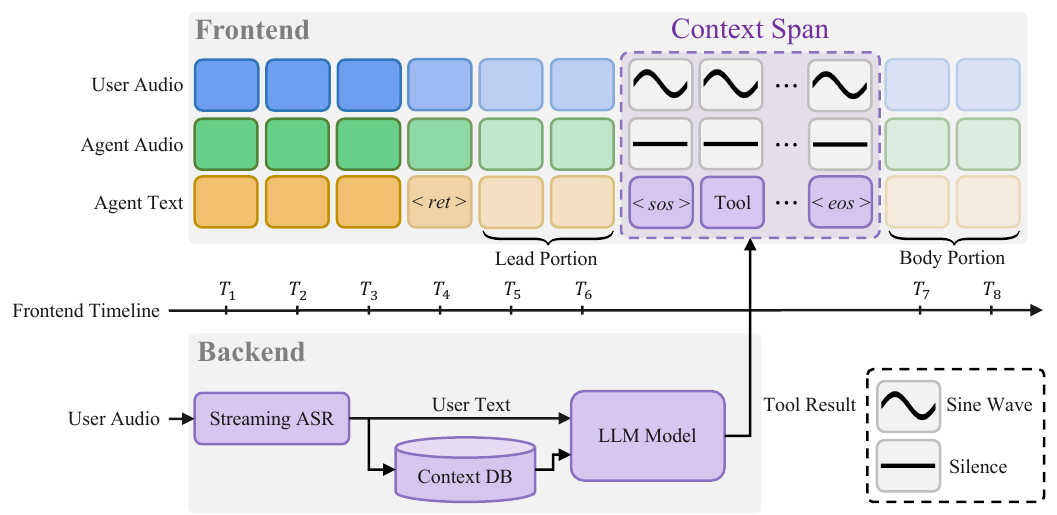}
\caption{Overall architecture of proposed method. Backend  works asynchronously, and results are injected as Context Span. 
}
\label{fig:context_span}
\end{figure*}

\section{Related Works}
\label{sec:related}

Full-duplex speech models \cite{defossez2024moshi,roy2026personaplex,zhang2026duplexsla} have been researched with various objectives.
Moshi\cite{defossez2024moshi} modeled the speech of both speakers and the agent's inner monologue text as multiple streams on a single timeline, so that turn-taking emerges from the model's own predictions rather than from an external module. PersonaPlex\cite{roy2026personaplex} built upon Moshi by injecting voice prompts and role prompts to add voice and role control. DuplexSLA\cite{zhang2026duplexsla} extends this line of work with an action-token-based approach that enables thinking before responding, moving toward a foundation model capable of taking actions. All of these provide excellent speech-to-speech baselines. However, while they focus on mastering conversational skills, these models still require the real-time information gathering capabilities of LLMs.

To resolve this, various models have attempted to bridge LLMs and full-duplex models\cite{chien2026moshirag,arora2025stream,kuroki2026kame}. 
StreamRAG\cite{arora2025stream} enables low-latency retrieval by proactively generating text queries during streaming speech before a user turn ends. However, it is primarily designed for predictive tool execution over static text databases rather than handling true full-duplex conversational interactions.
KAME\cite{kuroki2026kame} achieves full-duplex interaction by processing continuous audio streams, but it periodically triggers LLM calls at fixed time intervals regardless of dialogue context, leading to redundant execution and severe computational waste.
MoshiRAG\cite{chien2026moshirag} improved the factual correctness of full-duplex models using retrieval trigger tokens and an asynchronous backend. They claimed that they had shown the first full-duplex voice model with RAG. However, because it compresses text information and inserts it into the user audio token, it is difficult to deliver precise data like exact time or big number. Injecting external information into the user stream is also risky. The model may mistake it for the user's actual utterance, even if it is considered during training. Since the model should read all user audio tokens that are affected by injection, it may consume time to get all information due to long vector lengths. To address these limitations, we propose a novel approach named \textbf{Context Spanning}. 



\section{Architecture}
\label{sec:arch}

\subsection{Pipeline}
\label{ssec:pipeline}

 The overall pipeline is illustrated in Fig.~\ref{fig:context_span}. The retrieval system, data generation pipeline, and backend system follow the methodology of MoshiRAG~\cite{chien2026moshirag}. The frontend model performs real-time duplex conversation inference using agent text, agent audio, and user audio inputs, integrated with MoshiRAG's retrieval token-based backend. The Context Database (DB) is initialized to include optional user metadata including location, and timezone, etc.
 The user's audio is transcribed by real-time Automatic Speech Recognition (ASR) and accumulated in the Context DB. When a retrieval token is predicted, the output is computed by the LLM model based on the Context DB. The frontend model first generates the lead portion that does not depend on external information, and then produces the body portion containing reference-grounded content once the external knowledge arrives. With this pipeline, once the backend system completes an action and returns a result, the result needs to be passed to the frontend model in time. We use Context Spanning for this, that directly injects token frames into a duplex speech model architecture.

\begin{figure*}[t]
\centering
\includegraphics[width=0.80\textwidth]{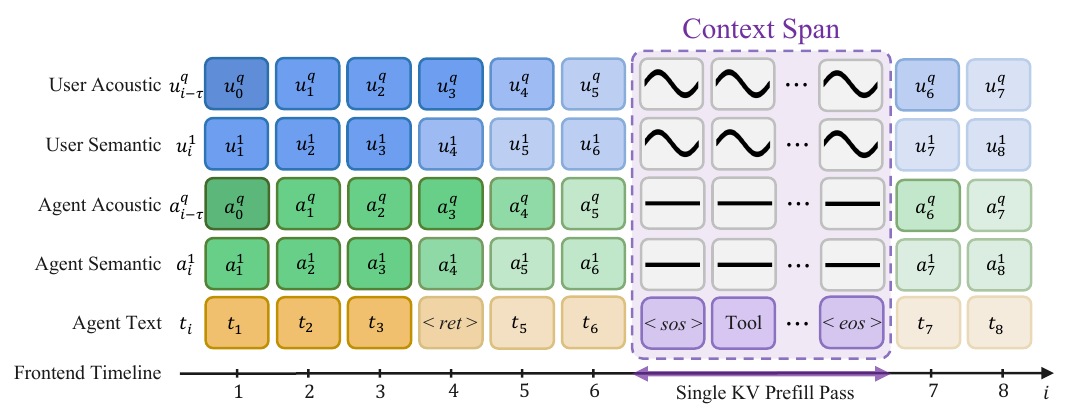}
\caption{Context Spanning on the Moshi architecture with an acoustic delay of $\tau = 1$. $u^{1}_{i}$ and $u^{q}_{i}$ denote user semantic and user acoustic tokens, $a^{1}_{i}$ and $a^{q}_{i}$ denote agent semantic and agent acoustic tokens, and $t_{i}$ denotes agent text tokens at timestep $i$, where $q \in \{2, \dots, 8\}$.}
\label{fig:span_grid}
\end{figure*}


\subsection{Context Spanning}
\label{ssec:cs}

In MoshiRAG, an ablation study demonstrated that injection-based methods can be more effective when handling external data. PersonaPlex also adopts a structure that injects voice prompts and system prompts, showing strong metrics in Service-Duplex-Bench experiments based on pre-injected information. However, all of these experiments were conducted exclusively by pre-injecting frames prior to the conversation, while frame injection directly into the stream during an ongoing conversation remains  unexplored.

When real-time information is required upon the generation of a retrieval token \texttt{<ret>}, an asynchronous backend executes RAG, MCP, and tool calls, injecting the search results within a span enclosed by \texttt{<sos>} and \texttt{<eos>}, which mean \textit{start of span} token and \textit{end of span} token, respectively. In the SentencePiece tokenizer we use, \texttt{<ret>}, \texttt{<sos>}, and \texttt{<eos>} are assigned token IDs 4, 12, and 13 respectively, all functioning as byte-fallback tokens originally.  In the span, agent voice is set to silence and the user audio input is set to a 440 Hz sine wave as in PersonaPlex's system prompt. 

Span injection does not require autoregressive sampling. The span tokens are given directly rather than sampled, so the model only processes them to update its KV cache for subsequent decoding. Since the Moshi architecture is built entirely from causal Transformers, span tokens can be prefilled in a single forward pass\cite{pope2023efficiently}. For a given integer $n$, assuming identical position embeddings, processing a sequence of $n$ positions in a single chunk yields attention outputs identical to those of $n$ sequential per-token steps due to causal masking. This equivalence holds across both temporal and depth transformer in Moshi architecture. In the training phase, we exclude this injected span from the loss function.

 Fig.~\ref{fig:span_grid} shows an example of Context Spanning in the Moshi token frame. The Moshi architecture contains a mechanism called acoustic delay. Acoustic tokens are input and predicted $\tau$ frames later than the semantic token of the same frame. However, applying acoustic delay directly to the span frames complicates the prefill forward pass due to frame misalignment. Therefore, during training data preparation, we inject the span after the delay is applied to the vectorized frames. During inference, this is achieved simply by inserting the span frames directly into the stream where the acoustic delay has already been applied. 


\section{Experiments and Results}
\label{sec:exp}

\subsection{Dataset and Training}

We used the Natural Questions~\cite{kwiatkowski2019natural}, HotpotQA~\cite{yang2018hotpotqa}, and TriviaQA~\cite{joshi2017triviaqa} datasets, following the MoshiRAG pipeline. For tool calling and MCP capabilities, we used same prompts to make scripts, but with different dataset. We constructed datasets from text benchmarks, incorporating 125 tools from MCPToolBench++~\cite{fan2025mcptoolbench++} (87) and Google SGD~\cite{lee2022sgd} (38). We excluded 65 tasks from MCPToolBench++ that are unsuitable for voice assistants (e.g., Web Browser, Filesystem, and Map Geocode). SODA~\cite{kim2023soda} was included for everyday conversation. Dialogue placement via the Candor Corpus~\cite{reece2023candor} was applied to train natural turn-taking and backchanneling.

Scripts were generated by Gemma 4 31B\cite{team2024gemma}. They were synthesized using Fish Audio\cite{fish-speech-v1.4} with 5,164 single-speaker GLOBE\cite{wang2024globe} prompts filtered by UTMOSv2\cite{baba2024utmosv2} to ensure a predicted MOS above 3.2. The full 2,100-hour stereo corpus contains 195,257 dialogs, averaging 38.2s and 9.9 spoken turns each. This comprises everyday conversations (1,453h/102k), knowledge retrieval (389h/71k), tool use (195h/12k), and abstaining scenarios (63h/10k). Among the dialogues that involve retrieval, there is an average of 1.53 retrieval tokens per dialogue. \ret \ comes at the head of a sentence that needs external information, and span contents are injected after a delay sampled from $\mathcal{U}(0.6, 2.0)$ seconds. Span contents were generated by Gemma 4-26B-A4B, which is also used in the backend.

Moshi\cite{defossez2024moshi} handles numbers by splitting them into single digits and using byte-backoff. However, we observed that this strategy often leads to frame misalignment in causal inference, particularly with large numbers. So we fully verbalized all inputs using the NeMo text normalization model\cite{nemo} prior to tokenization.




We finetune all trainable parameters initialized from the public PersonaPlex-7B checkpoint to enable voice prompting. Optimization is performed using AdamW \cite{loshchilov2017decoupled} with a context length of 3,000 frames. The learning rates are set to 2e-6 for the temporal transformer and 4e-6 for the depth transformer. Training was conducted on two NVIDIA RTX Pro 6000 GPUs, and one full epoch of training took 8 hours. For evaluation, we used two NVIDIA RTX Pro 6000 GPUs, dedicating one to the frontend speech model and the other to the retrieval backend. For the retrieval backend in experiments, we used the Gemma 4-26B-A4B and GPT-4.1. Unless otherwise specified, all experiments used the Gemma 4-26B-A4B backend.

\subsection{Latency Analysis}
\label{ssec:latency}

Moshi's Mimi encoder operates at $12.5\,\text{Hz}$, producing one frame every $80\,\text{ms}$. Once the backend computation is finalized, we wait until an $80\,\text{ms}$ window is secured. Our model should complete both the Context Span prefill and one step of autoregressive decoding within this $80\,\text{ms}$ budget. In Table~\ref{tab:latency}, detailed metrics including the mean, standard deviation, and 99th percentile are reported. The latency increases sub-linearly with longer spans. These results show that Context Spanning enables encoding rich information in real-time.
\begin{table}[t]
\centering
\caption{Context Span processing latency. $P_{99}$ denotes 99th percentile. $n=0$ means no span, same as vanilla Moshi.}
\label{tab:latency}
\footnotesize
\resizebox{\columnwidth}{!}{%
\begin{tabular}{@{}lccccc@{}}
\hline
$n$ (frames)  & 0 & 16 & 64 & 256 & 600 \\
\hline
Latency ($\text{ms}$)& $34.7 \pm 0.4$ & $56.1 \pm 0.2$ & $57.5 \pm 0.2$ & $62.0 \pm 0.3$ & $86.1 \pm 0.3$ \\
$P_{99}$ ($\text{ms}$) & $35.2$     & $56.3$         & $57.7$         & $62.3$         & $86.3$         \\
\hline
\end{tabular}%
}
\end{table}

\begin{table*}[!t]
\centering
\caption{QA Benchmarks. `ref.' denotes the accuracy \% that correct reference document is provided; `resp.' denotes the accuracy \% that model response correctly. \underline{Underlined} model metrics are reprinted from MoshiRAG, `-' denotes values inaccessible.}
\label{tab:qa_math}
\setlength{\tabcolsep}{3pt}
\resizebox{\textwidth}{!}{%
\begin{tabular}{@{}l*{8}{c}|*{10}{c}@{}}
\hline
 & \multicolumn{8}{c}{Spoken QA} & \multicolumn{10}{c}{Mathematical Reasoning} \\
\cline{2-9}\cline{10-19}
 & \multicolumn{2}{c}{LlamaQ} & \multicolumn{2}{c}{WebQ} & \multicolumn{2}{c}{TriviaQA} & \multicolumn{2}{c}{HaluEval}
 & \multicolumn{2}{c}{AddSub} & \multicolumn{2}{c}{MultiArith} & \multicolumn{2}{c}{SinglEq} & \multicolumn{2}{c}{SVAMP} & \multicolumn{2}{c}{GSM8K} \\
\cline{2-3}\cline{4-5}\cline{6-7}\cline{8-9}\cline{10-11}\cline{12-13}\cline{14-15}\cline{16-17}\cline{18-19}
Model & ref. & resp. & ref. & resp. & ref. & resp. & ref. & resp.
 & ref. & resp. & ref. & resp. & ref. & resp. & ref. & resp. & ref. & resp. \\
\hline
\underline{GLM-4-Voice}
 &  & 64.7 &  & 32.2 &  & 39.1 &  & 21.2
 &  & 59.4 &  & 62.0 &  & 71.0 &  & 4.0 &  & 29.0 \\
\underline{STITCH-S}
 &  & 73.3 &  & 50.2 &  & 50.0 &  & --
 &  & \textbf{81.7} &  & 87.9 &  & \textbf{91.7} &  & 72.2 &  & 56.7 \\
\underline{MoshiRAG$_{\text{Gemma3}}$}
 & 83.0 & 80.3 & 71.5 & 67.2 & 73.7 & 69.6 & 42.0 & 36.3
 & 76.6 & 61.7 & 87.1 & 69.0 & 83.2 & 68.2 & 74.1 & 55.0 & 66.2 & 33.9 \\
\underline{MoshiRAG$_{\text{GPT-4.1}}$}
 & 87.8 & 80.6 & 77.7 & \textbf{68.9} & 86.8 & 78.2 & 61.2 & 51.3
 & 87.9 & 64.8 & 87.1 & 76.0 & 89.6 & 72.9 & 80.5 & 61.1 & 70.8 & 43.2 \\
\underline{MoshiRAG$_{\text{Tavily}}$}
 & 84.6 & 78.2 & 73.5 & 66.1 & 84.9 & 77.5 & 54.3 & 47.0
 &  & -- &  & -- &  & -- &  & -- &  & -- \\
\underline{Vanilla Moshi}
 &  & 62.3 &  & 26.6 &  & 22.8 &  & 10.5
 &  & 8.3 &  & 9.8 &  & 18.4 &  & 9.7 &  & 2.1 \\
Ours$_{\text{Gemma4}}$
 & 85.3 & 81.7 & 67.7 & 59.1 & 71.5 & 68.9 & 39.6 & 33.7
 & 78.5 & 76.2 & 94.5 & \textbf{89.7} & 82.1 & 77.6 & 86.0 & 81.1 & 70.4 & 62.5 \\
Ours$_{\text{GPT-4.1}}$
 & 89.9 & \textbf{83.3} & 79.4 & 66.7 & 91.1 & \textbf{83.8} & 68.3 & \textbf{55.5}
 & 82.6 & 74.7 & 95.2 & \textbf{89.7} & 85.9 & 82.2 & 89.2 & \textbf{85.4} & 75.7 & \textbf{67.4} \\
\hline
\end{tabular}%
}
\end{table*}

\subsection{QA Benchmarks}
Following MoshiRAG, pre-computed GPT answers were injected after a fixed delay. While MoshiRAG used a 1.5-second delay, GPT-4.1 actually took 0.77s for average response time in benchmarks, so we applied a 0.8-second delay in our experiments. We used real-time retrieval for the Gemma backend. We waited $0.5\,\text{s}$ after the \ret token for stable ASR as in MoshiRAG. We used the Qwen3-ASR-1.7B\cite{Qwen3-ASR} model for ASR. The retrieval delay has a mean of $1.08\,\text{s}$ and a standard deviation of $0.41\,\text{s}$. We tested our model on the Spoken QA and Math reasoning datasets, where the math domain was  unseen during training. As shown in Table~\ref{tab:qa_math}, our model performs comparably to MoshiRAG baselines on QA tasks. Especially on math datasets, our model demonstrates a superior ability to yield correct responses when provided with reference documents, outperforming MoshiRAG across response benchmarks. 

\begin{table}[t]
\centering
\caption{FullDuplexBench v1 results. \underline{Underlined} models are reprinted from the PersonaPlex paper. PPlex denotes PersonaPlex, Gemini denotes Gemini Live 2.5.}
\label{tab:fdb_v1}
\setlength{\tabcolsep}{3pt}
\resizebox{\columnwidth}{!}{%
\begin{tabular}{@{}llcccc@{}}
\hline
Task & Metric & Ours & \underline{PPlex} & \underline{Moshi} & \underline{Gemini} \\

\hline
\multirow{2}{*}{Turn-Taking} & TOR ($\uparrow$) & 0.899 & 0.992 & 0.941 & 0.655 \\
 & Latency, s ($\downarrow$) & 0.064 & 0.070 & 0.265 & 1.301 \\
 \hline
\multirow{2}{*}{Pause Handling} & TOR, Candor ($\downarrow$) & 0.824 & 0.662 & 0.980 & 0.310 \\
 & TOR, synthetic ($\downarrow$) & 0.861 & 0.584 & 0.985 & 0.255 \\
 \hline
\multirow{3}{*}{Backchannel} & TOR ($\downarrow$) & 0.782 & 0.327 & 1.000 & 0.091 \\
 & Frequency, per s ($\uparrow$) & 0.150 & 0.025 & 0.001 & 0.012 \\
 & JSD ($\downarrow$) & 0.716 & 0.649 & 0.957 & 0.896 \\
 \hline

\multirow{3}{*}{User Interruption} & TOR ($\uparrow$) & 0.925 & 1.000 & 1.000 & 0.891 \\
 & GPT-4o rating, 0--5 ($\uparrow$) & 3.924 & 4.210 & 0.765 & 3.376 \\
 & Latency, s ($\downarrow$) & 0.598 & 0.400 & 0.257 & 1.183 \\
\hline
\end{tabular}%
}
\end{table}

\subsection{Full Duplex Benchmarks}
Full Duplex Benchmarks \cite{lin2025full,lin2026full-v3} were proposed to evaluate performance based on special capabilities required as a speech-to-speech model rather than a cascade speech model. In Table~\ref{tab:fdb_v1}, we evaluate our model by FullDuplexBench v1. TOR is the take-over rate, the ratio of clips in which the model takes the turn. Backchannel frequency is the number of backchannels per second, JSD is the Jensen-Shannon divergence from the human backchannel timing distribution, latencies are in seconds, and the interruption response is rated by GPT-4o. Turn-Taking Latency showed strong performance. Although backchannel frequency increased significantly, instances recognized as TOR also grew. While Candor-based backchannel generation effectively raised frequency, better handling of turn-taking should be considered. The model showed lower performance in Interruption and Pause Handling. 

In Table~\ref{tab:fdb_v3}, we evaluate our model's tool-calling performance and turn-taking dynamics using FullDuplexBench v3. MoshiRAG is the released MoshiRAG checkpoint with the Gemma 3-27B reference LLM using the same tool router and prompt as FullDuplexBench v3. Tool selection, argument accuracy, response quality, and pass rate are fractions out of the 100 scenarios. Take-turn, interruption, and filler rates are percentages, and latency is the task-completion time in seconds. The pass rate demonstrates our model's competitiveness as a tool-calling model. But since fillers are injected at the lead portion in our model and MoshiRAG, it degrades overall metrics, especially in Filler rate. Furthermore, the model shows weakness in multi-turn connected conversations.

\begin{table}[t]
\centering
\caption{FullDuplexBench v3 results. \underline{Underlined} systems are reprinted from the original benchmark paper. GPT denotes GPT-Realtime, Gemini denotes Gemini Live 3.1. }
\label{tab:fdb_v3}
\setlength{\tabcolsep}{5pt}
\resizebox{\columnwidth}{!}{%
\begin{tabular}{@{}llcccc@{}}
\hline
Task & Metric & Ours & MoshiRAG & \underline{GPT} & \underline{Gemini} \\
\hline
\multirow{4}{*}{Tool Use}
 & Tool selection ($\uparrow$) & 0.855 & 0.738 & 0.876 & 0.817 \\
 & Argument accuracy ($\uparrow$) & 0.567 & 0.440 & 0.680 & 0.588 \\
 & Response quality ($\uparrow$) & 0.411 & 0.255 & 0.792 & 0.718 \\
 & Pass rate ($\uparrow$) & 0.470 & 0.280 & 0.600 & 0.540 \\
\hline
\multirow{4}{*}{\begin{tabular}[c]{@{}l@{}}Turn-Taking\\ Dynamics\end{tabular}}
 & Take-turn rate, \% ($\uparrow$) & 95.0 & 94.0 & 96.0 & 78.0 \\
 & Latency, s ($\downarrow$) & 5.83 & 7.69 & 6.89 & 4.25 \\
 & Interruption rate, \% ($\downarrow$) & 73.7 & 58.5 & 13.5 & 19.2 \\
 & Filler rate, \% ($\downarrow$) & 96.0 & 92.3 & 16.9 & 31.7 \\
\hline
\end{tabular}%
}
\end{table}
\section{Limitation and Future Work}
\label{sec:limits}

Since token injection uses one frame per token, Context Spanning requires a high memory overhead during dialogue.  Applying our methods to models like DuplexSLA \cite{zhang2026duplexsla}, which optimize multiple text tokens in a single frame, could be considered as future work.

The role of injected context span is currently limited. In our setup, context span serves as text only signals alongside agent text tokens. All audio tokens are provided as fixed indicators, leaving room to inject more information within the same frame size. Additionally, tokens may carry emotional tone or system instructions. Expanding text tokens to express diverse behaviors, or developing new architectures centered on 'Thinking and Talking' mechanisms by context control with span, remains a promising direction.

 We constructed all of our data with a TTS model. For more clarity in natural conversation, we should use real conversation datasets with voice prompts in train data, but they are not accessible at this time. In addition, reasoning performance still relies on the backend system including ASR and the LLM.  Our system focuses on injecting LLM-retrieved information rather than acting as a generalized agent. Standardizing these behavioral patterns into a unified architecture will be a key next step for duplex speech foundation models.



\section{Conclusion}
\label{sec:conc}

In this paper, we propose Context Spanning, a framework that connects a full-duplex speech model with an external LLM backend. During real-time conversation, the system inserts retrieval results directly into the ongoing duplex speech stream. Our approach can preserve information without losing any detail. Benchmark evaluations on full-duplex interaction and spoken question answering show that our model achieves practical performance in real-time information retrieval and tool execution. However, handling complex real-world conversations still requires further refinement. Future work includes improving context memory management, reducing reliance on backend ASR, adding explicit reasoning capabilities, applying reinforcement learning, and expanding the framework to support multiple languages. We expect that applying Context Spanning to existing duplex models will be a promising direction for the duplex speech model research community.

\section{Acknowledgment}
Generative AI tools were used for language editing only. All technical content
and conclusions are the work of the authors.

\bibliographystyle{IEEEbib}
\bibliography{refs}

\end{document}